\pdfoutput=1

\documentclass[11pt]{article}

\usepackage[]{ACL}

\usepackage{times}
\usepackage{latexsym}

\usepackage[T1]{fontenc}

\usepackage[utf8]{inputenc}

\usepackage{microtype}

\usepackage{xspace}
\usepackage[inkscapearea=page]{svg}
\usepackage{subcaption}
\usepackage{multirow}

\usepackage{multicol}
\usepackage{amsmath,amssymb}
\usepackage{bbm}

\usepackage{enumitem}
\usepackage{csquotes}
\usepackage{booktabs} 
\usepackage{amsthm}

\title{Combining Confidence Elicitation and Sample-based Methods for Uncertainty Quantification in Misinformation Mitigation}

\author{Mauricio Rivera\textsuperscript{1}, Jean-Fran\c{c}ois Godbout\textsuperscript{2}, \\ \textbf{Reihaneh Rabbany\textsuperscript{1},  Kellin Pelrine\textsuperscript{1}} \\
   \textsuperscript{1}McGill University; Mila \quad
   \textsuperscript{2}Universit\'e de Montr\'eal \quad
}

\begin{document}
\maketitle
\begin{abstract}

Large Language Models have emerged as prime candidates to tackle misinformation mitigation. However, existing approaches struggle with hallucinations and overconfident predictions. We propose an uncertainty quantification framework that leverages both direct confidence elicitation and sampled-based consistency methods to provide better calibration for NLP misinformation mitigation solutions. We first investigate the calibration of sample-based consistency methods that exploit distinct features of consistency across sample sizes and stochastic levels. Next, we evaluate the performance and distributional shift of a robust numeric verbalization prompt across single vs. two-step confidence elicitation procedure. We also compare the performance of the same prompt with different versions of GPT and different numerical scales. Finally, we combine the sample-based consistency and verbalized methods to propose a hybrid framework that yields a better uncertainty estimation for GPT models. Overall, our work proposes novel uncertainty quantification methods that will improve the reliability of Large Language Models in misinformation mitigation applications.

\end{abstract}

\section{Introduction}
\label{sec:intro}

It has become crucial to combat the spread of misinformation and detect deceptive content on social media. Misinformation can challenge the fairness of elections \citep{meel2020fake}, perpetuate a cascade of rumors resulting in significant financial losses \citep{marcelo2023fact} and even endanger lives \citep{loomba2021measuring}. Recent work has demonstrated that Large Language Models (LLMS) can be prime candidates for countering misinformation \citep{pelrine2023towards, flores22adversarial, kaliyar2021fakebert, pelrine2021surprising}. However, their usage in high-value applications is held back by the \textit{hallucination} problem. The best LLMs have been trained to produce convincing responses, thus they often appear overconfident \cite{ji2023survey}. Such combination creates instances where the models yield answers that, while sounding reasonable, are significantly inaccurate. Hence, because low uncertainty---or high confidence---does not guarantee accuracy \citep{huang2023look}, it is essential to develop methods to estimate the levels of uncertainty of these models.

Furthermore, since closed-source LLMs, such as GPT-3.5 and GPT-4, often do not provide access to the model logits or embeddings to evaluate their reliability, there is also a need for \textit{non-logit-based} uncertainty quantification methods. In this paper, we propose a framework that combines verbalized confidence methods, which verbally convey information about the model's intrinsic uncertainty, with sample-based methods, which distills an estimation of the model's certainty through the consistency of its answers. This approach allows us to derive a hybrid uncertainty score that provides better model calibration on the LIAR dataset, a commonly used repertory of short fake-news statements \citep{wang2017liar}.

We compare the performance of different sample-based consistency methods across various temperature levels and sample sizes. Specifically, we compare several 
 known methods: self-consistency \citep{wang2022self}, an adaptation of selfcheckGPT \citep{manakul2023selfcheckgpt}; the normalized standard deviations;  and the range of predicted class probabilities. We also develop two methods, named SampleAvgDev and Deviation-Sum, and compare their performance with the other sample-based methods. In addition, we explore the distributional and performance shifts of single-step vs. two-step confidence elicitation, showing that the two-step confidence elicitation provides the best calibration. We also carry out comprehensive experiments to evaluate our prompting strategies, including a comparison of the performance of the explain-score prompt \citep{pelrine2023towards} on different truth scales and various versions of GPT. Finally, we integrate all of the above results to the BSDetector framework \citep{chen2023quantifying}, which allows us to evaluate the models' uncertainty. 

Overall, our key contributions are the following:
\begin{itemize}
    \item We  compare the calibration capabilities of various sample-based consistency methods in the context of misinformation mitigation and report how their performances scale with temperature and sample size.
    \item We implement an adapted version of \citet{chen2023quantifying}'s BSDetector framework that leverages the synergy between sample-based consistency and confidence elicitation methods. As a result, all proposed methods exhibit enhanced performance, achieving an ECE score lower than 0.13, which outperforms previous misinformation mitigation calibration solutions on the LIAR dataset \cite{pelrine2023towards}.
    \item We propose the SampleAvgDev sample-based consistency method paired with a two-step confidence elicitation prompt and conclude that this approach is the most efficient calibration technique for our model with an ECE score of 0.076.
\end{itemize}





\section{Background and Related Work}
\label{sec:background}

\subsection{Misinformation Detection}
There are several misinformation detection solutions, which can be categorized into content-based and network-based approaches \citep{shu2017fake}. Content-based approaches, the focus of this paper, tackle this issue by analyzing the text, images, or multimedia elements of a message to determine its veracity. While prior solutions' generalization abilities have been limited \citep{sharma2019combating}, GPT-4 has emerged as the top candidate for misinformation detection and classification \cite{pelrine2023towards, quelle2023perils} by demonstrating superior performance on various misinformation datasets. Still, its overall performance and reliability are not robust enough for direct real-world application, as confirmed by \cite{pelrine2023towards}, which highlight that models often hallucinate and are overconfident in their responses.

\subsection{Uncertainty Quantification}
Uncertainty quantification methods, which attempt to measure the uncertainty level of model outputs, remain one of the most effective risk assessment methods for Machine Learning models \cite{hullermeier2021aleatoric}.
In this paper, we focus on tackling \textit{epistemic} uncertainty, meaning the uncertainty coming from the LLM's parameters \cite{kendall2017uncertainties} by combining sample-based and verbalized confidence methods.

\subsubsection*{Verbalized Confidence Methods} Benefiting from GPT's impressive verbal capabilities, it is possible to directly elicit these LLM's uncertainty via verbal cues, such as those demonstrated by \citet{lin2022teaching} verbalized confidence approach. This technique improves the model's calibration \cite{tian2023just}. Verbalized confidence methods also benefit from prompt engineering principles, where leveraging Chain-of-Thought (CoT) prompting \citep{wei2022chain} improves the model's calibration and generates adequate reasoning processes \citep{xiong2023can}. 

\subsection{Sample-based Consistency Methods} Sample-based methods estimate uncertainty by leveraging the inherent stochasticity of LLMs. In our context, we can simulate stochastic answers by setting GPT's temperature parameter T > 0 \cite{huang2023look}. In general, this method involves generating multiple stochastic responses for the same question, and use the consistency among those answers to estimate the model's uncertainty. In sample-based evaluations, this approach has been shown to consistently outperform purely verbalized methods \citep{xiong2023can}; it has also achieved even better performance when combined with verbalized techniques in hybrid methods \citep{chen2023quantifying, xiong2023can}.  In the next section, we provide additional background on the theoretical basis of sample-based consistency methods used in literature.

\subsubsection*{Self-consistency} 
Self-consistency leverages the intuition that a complex reasoning problem accepts different ways of thinking leading to its unique correct answer. Consequently, this approach chooses the optimal answer by finding the most consistent answer \citep{wang2022self}. Interestingly, prior work has confirmed that self-consistency boosts the performance of chain-of-thought prompting and is robust to imperfect prompts \citep{wang2022self}.

\subsubsection*{SelfCheckGPT}
SelfCheckGPT capitalizes on the idea that the LLMs' hallucinations from stochastic sampled responses are more likely to diverge compared to well-known answers \cite{manakul2023selfcheckgpt}. Therefore, this sample-based consistency method measures the divergence of stochastic samples based on a non-stochastic reference answer. 

\subsubsection*{Normalized Standard Deviations} It has been shown that variance can serve as an indicator for uncertainty \citep{lin2023generating}. The idea here is simple: a sample with higher standard deviation indicates that the model is more uncertain of its prediction.

\section{Methodology}
\label{sec:methodology}

\subsection{Data}
All experiments were performed using the LIAR dataset \citet{wang2017liar}, one of the most widely-used benchmarks for fake news detection. It contains nearly 13K short political statements with veracity labeled on a 6-point scale, which we dichotomized by splitting the labels in the middle to get binary labels. This is a standard approach in the literature \cite{pelrine2023towards}.

\subsection{GPT Prompt Engineering}
Based on previous results \citep{pelrine2023towards}, we use the `Explain-Score' prompt on GPT4-0314 as the basis of all developed prompts. 

\begin{quote}
    \textbf{Explain-Score prompt}: “role”: “user”, “content”: ‘Rate the truthfulness
of the following statement: "STATEMENT" Your
task is to provide a score from 0 to 100, where
0 represents definitively false and 100 represents
definitively true, but you must not state your score
until you've presented a thorough analysis. Do
not begin your response with a number. First
write your analysis, then write a vertical bar "|",
then finally state your score.’
\end{quote}

The exact prompts are detailed in the Appendix \ref{sec:appendix_prompt}.

\subsubsection*{Truth scales prompts}
To test GPT's ability for hard-classification, we provide Politifact's truth-O-meter scale within the context of the prompt to get a direct 6-point classification truthfulness score. We denote this prompt as `Politifact'. In addition, we tested GPT's ability for categorical classification on a multiple-choice 3-point scale: i.e., A) True, B) False, C) Somewhat True/ Somewhat False. We denote this prompt as `3way-Categorical'.

\subsubsection*{Distributional Shift prompt}
To analyze GPT's distribution and performance shifts in single vs. 2-step confidence elicitation, we prompted the model using different procedures. For the single step confidence elicitation procedure, we request GPT to rate the truthfulness of a LIAR data statement while demanding to rate its uncertainty over that answer, all within a single prompt. We denote this prompt `single-step-uncertainty'. For the 2-step confidence elicitation procedure, we first obtain a truthfulness score and explanation from the GPT model using the Explain-Score prompt. Then, we prompt the model a second time, now requesting to rate the uncertainty of its previously generated truthfulness score and explanation for the given LIAR data statement. We denote this prompt as `2-Step-Uncertainty'. 

\subsubsection*{CoT Prompt}
Because (CoT) prompting is known to enhances the model's calibration and generates adequate reasoning processes \citep{xiong2023can}, we devised a prompt inspired from the Explain-Score approach where we specify GPT to generate a truthfulness score paired with a CoT-format explanation. We denote this prompt as `CoT-Explain-Score'.

\subsection{Sample-based consistency methods}
In this section, we describe the sample-based consistency methods used in our experiments. For these methods, we generate k-stochastic outputs from the same prompt. We denote the stochastic generated answers \(a_i\) from a fixed answer set, \(a_i \in A\), where \(i = 1, \ldots, k\) indexes the i-th sample. The answer set corresponds to the truthfulness or uncertainty scale used in the prompt. For most of the experiments, A = [0-100]. For selfCheckGPT, we additionally consider an non-stochastic reference answer to be \(a_r\) generated by setting the parameter T = 0. It is important to note that all sample-based consistency methods were min-max normalized to obtain a common 0-1 uncertainty score.

\subsubsection*{Self-consistency} Note that we attempted to adapt the Self-consistency framework to our context \cite{wang2022self}. Specifically, the self-consistency score corresponds to the most frequent score from the k-stochastic answers weighted by \(\frac{1}{k}\). It is computed as follows:
\[a^* = \arg\max_a \frac{1}{k}\sum_{i=1}^k \mathbbm{1}(a_i = a)\] 

\subsubsection*{SelfCheckGPT} We also attempted to adapt the SelfCheckGPT framework to our context \cite{manakul2023selfcheckgpt}. Specifically, the selfCheckGPT score is an average of the amount of stochastic answers that match the non-stochastic reference answer. It is computed as follows:

\[a^* = \frac{1}{k}\ \sum_{i=1}^k \mathbbm{1}(a_i = a_r)\].

\subsubsection*{Sample average deviation} The sample average deviation (SampleAvgDev) calculates the average of the absolute difference between the i-th stochastic answer and the halfpoint of our classification (50, in our case).
The rationale behind this method is rooted in our prompt structure: Given that we instruct GPT models to assess the truthfulness of a statement on a 0-100 scale, here 0 represents definitely false and 100 represents definitely true, we can capture the model's uncertainty by measuring the deviation of its prediction from the halfpoint of our classification (50). Furthermore, averaging these deviations from the halfpoint aims to provide a better representation of the model's actual uncertainty by the law of large numbers, hence leveraging the principle of consistency for uncertainty quantification. Specifically, the SampleAvgDev score is computed as follows:

\[a^* = \frac{1}{k}\sum_{i=1}^k|a_i - 50|\] 

\subsubsection*{Normalized standard deviations} The Normalized standard deviation (Norm. std) method involves taking the standard deviation of k-stochastic answers.

\subsubsection*{Deviation-Sum} Deviation-Sum was developed to estimate the model's uncertainty via the total absolute spread of the stochastic answers according to their mean. Namely, letting \(\bar{a_k}\) denoting the k-sample average, the Deviation-Sum's answer is computed as follows: 

\[a^* = \sum_{i=1}^k |\bar{a_k}\ - a_i|\]

\subsubsection*{Predicted class probability margin} The predicted class probability range (PredClassMargin) computes the margin between the most frequent and the least frequent score. Intuitively, a wider range implies higher uncertainty, and is particularly relevant to multiclass classification tasks. 

\subsection{Evaluation Metrics}

\subsubsection*{Expected calibration error (ECE) } This metric is commonly used to evaluate model calibration \cite{tian2023just, guo2017calibration}. First, we separate the model's predictions into bins \(B_i\) with quantile scaling, i.e., each bin is scaled to have the same number of examples, where \(i = 1, \ldots, m\) indexes the \(m\) bins (we use m = 10). Then, we measure the average accuracy \(acc(B_i)\) and average uncertainty \(uncert(B_i)\) of each bin. Finally, we compute the sum of absolute differences between the average accuracies and uncertainties, weighted by the number of samples \(n\) within each bin. A lower ECE implies a better model calibration. Explicitly, the ECE is computed as follows: 

\[ECE = \sum_{i=i}^m \frac{|B_i|}{n}\ |acc(B_i) - uncert(B_i)|\].

\subsubsection*{Brier Score} In a broad sense, the Brier Score is a score function that measures the accuracy of probabilistic predictions. A lower Brier score indicates better model calibration. Consider a binary training example \(x_i\) and its true binary label \(y_i\), where \(i=1, \ldots, n\). Then, the Brier is computed as follows:

\[Brier  Score = \frac{1}{N}\ \sum_{i = 1}^N (uncert(x_i) - \mathbbm{1}(x_i = y_i))^2\]. 

\subsubsection*{Kolmogorov-Smirnov test} The Kolmogorov-Smirnov (K-S) test is a nonparametric test used to test whether two samples come from the same distribution as an hypothesis test. The null hypothesis, which states that the two samples come from the same distribution, is rejected if the p-value generated from this test is smaller than the significance threshold. In our case, we use a significance value of 0.05.

\subsubsection*{Not Numbers} In some instances, the GPT models refused to give a numerical score of the LIAR's statements truthfulness. Hence, we denoted such occurrences as 'Not Numbers' (N.Ns) answers. 

All other evaluation metrics in our analysis are specified in Appendix \ref{sec:appendix_metrics}.

\section{Experiments}
\label{sec:experiments}

\subsection{Sample-based Consistency Methods}
In general, sample-based methods for uncertainty quantification generate an estimation of the model's uncertainty through the consistency of its answers. In our case, we first generate k-stochastic samples of truthfulness scores from the Explain-Score prompt.
Then, we use those k-samples as input to a sample-based consistency method, which in turn produces a 0-100 uncertainty score. Reminiscent to selecting a summary statistic in Bayesian analysis, we posit that the choice of the sample-based consistency method reflects distinct characteristics of the sample's distributions. For instance, self-consistency reflects the mode of the distribution, while Norm-std and the predicted class probability range contains information about the sample's spread. 

\begin{table}[!ht]
\caption{\textbf{Sample-based Consistency Methods}}
    \centering
    \begin{tabular}{lccc} 
         \textbf{Method} &  \textbf{ECE} & \textbf{Brier Score} \\
         \midrule     
         self-consistency&  0.226& 0.303\\ 
         selfcheckGPT&  0.179& 0.354\\ 
         PredClassMargin&  0.267& 0.301\\ 
         \textbf{SampleAvgDev}& \textbf{0.139}& \textbf{0.291}\\
         Norm. std&  0.361& 0.421\\ 
         Deviation-Sum&  0.376& 0.423\\ 
         \bottomrule
\label{tab1}
    \end{tabular}
\end{table}

In Table \ref{tab1}, we show the model's binary-classification calibration performance across the previously discussed sample-based consistency methods for 10 samples and T = 1.0 on the Explain-Score prompt. 
The results confirm that SampleAvgDev outperforms all other methods, while Norm. std and Deviation-Sum have significantly lower performances. This is expected, for the distribution of the uncertainty scores of these methods are skewed to lower values (see Appendix \ref{sec: appendix_uncertainty_distr} for a visualization of this effect).

\subsection{Effect of sample size}

\begin{table*}[!ht]
\caption{\textbf{Sample-size effect}}
    \centering
    \begin{tabular}{lcccccc}
         \textbf{Method}&  \multicolumn{3}{c}{\textbf{ECE}}&  \multicolumn{3}{c}{\textbf{Brier Score}}\\
         \midrule
         \textbf{Sample size}& k=2&  k=5&  k=10&  k=2&  k=5& k=10\\
         \midrule
         self-consistency&  0.246 &  0.232 & 0.226 &  0.333 &  0.315 & 0.303 \\
         selfCheckGPT& 0.354 & 0.281 & 0.179 & 0.3496 &  0.356 & 0.354 \\
         PredClassMargin& 0.466 & 0.33 & 0.267 & \textbf{0.310} & 0.294 & 0.301 \\
         \textbf{SampleAvgDev} & \textbf{0.136} & \textbf{0.138} & \textbf{0.139} & 0.341 & \textbf{0.293} & \textbf{0.291}\\
         Norm. std& 0.428 & 0.404 & 0.361  & 0.414 & 0.425 & 0.421 \\
         Deviation-Sum& 0.415 & 0.389 & 0.376 & 0.415 & 0.429 & 0.423 \\
        \bottomrule
    \end{tabular}
    \label{tab2}
\end{table*}

\begin{table*} 
    \caption{\textbf{Temperature Ablation}}
    \centering
    \begin{tabular}{lcccccc}
         \textbf{Method}&  \multicolumn{3}{c}{\textbf{ECE}}&  \multicolumn{3}{c}{\textbf{Brier Score}}\\
         \midrule
         \textbf{Temperature}& T=0.0&  T=0.5&  T=1.0&  T=0.0&  T=0.5& T=1.0\\
         \midrule
         self-consistency&  0.269 &  0.226 & 0.226 &  0.333 &  0.326 & 0.303 \\
         selfCheckGPT& 0.170 & 0.152 & 0.179 & 0.348 &  0.337 & 0.354 \\
         PredClassMargin& 0.363 & 0.266 & 0.267 & 0.347 & 0.347 & 0.301 \\
         \textbf{SampleAvgDev} & \textbf{0.168} & \textbf{0.148} & \textbf{0.139}& \textbf{0.286} & \textbf{0.280} & \textbf{0.291}\\
         Norm. std& 0.439 & 0.425 & 0.361  & 0.420 & 0.401 & 0.421 \\
         Deviation-Sum& 0.458 & 0.448 & 0.376 & 0.456 & 0.424 & 0.423 \\
        \bottomrule
    \end{tabular}
    \label{tab3}
\end{table*}

\begin{figure}[!ht]
    \centering
    \caption{\textbf{Effect of Sample size on sample-based consistency methods}}
    \includegraphics[width=1\linewidth]{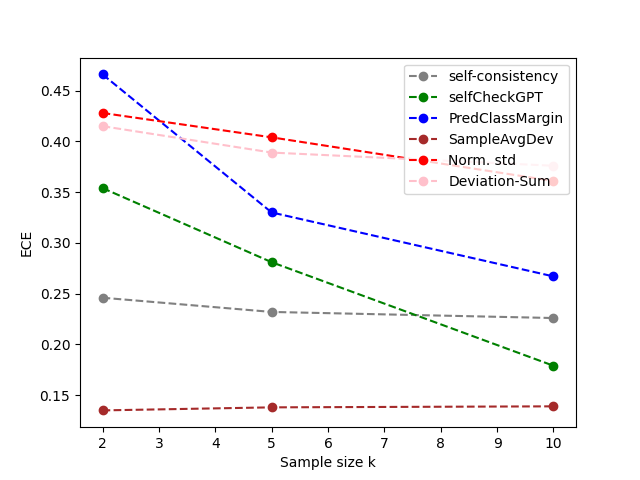}
    \label{fig1}
\end{figure}

We investigate the effect of varying sampling size on each proposed sample-based consistency method with the premise that the proposed methods might benefit from a larger sample-size. Indeed, previous work has proven that higher sample size leads to better uncertainty estimates in the BSDetector framework \citep{chen2023quantifying}. Table \ref{tab2} supports this claim, as nearly all proposed methods scale in calibration with sample size. Surprisingly, SampleAvgDev does not require a large sample size to have a performing ECE score. Yet, note the decrease in its Brier score suggests it also scales with sample size. Conversely, Figure \ref{fig1} displays that selfCheckGPT and PredClassMargin have nearly doubled their improvement in Expected Calibration Error (ECE), which suggests that the sample size significantly improves the efficacy of these methods.

\subsection{Temperature Ablation}
The influence of stochasticity on the suggested sample-based consistency methods could vary from one method to the next. In fact, reduced randomness inherently constrains the divergence of sample responses, thereby restricting the span of certain consistency methods. Consequently, we conduct a temperature ablation study on the Explain-Score prompt with 10 samples to examine the influence of stochasticity on each proposed methods, as shown in Table \ref{tab3}. 

\begin{figure}[!ht]
    \centering
    \caption{\textbf{Temperature ablation experiment on sample-based consistency methods}}
    \includegraphics[width=1\linewidth]{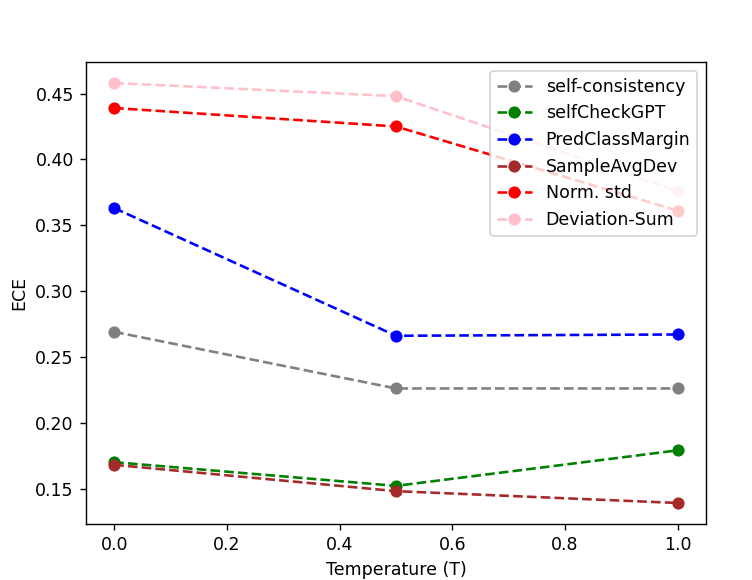}
    \label{fig2}
\end{figure}

Indeed, as illustrated in Figure \ref{fig2}, most methods show small improvements with temperature. We hypothesize that this effect is more pronounced in sample-based consistency methods that capitalize on a larger uncertainty score distribution spread, such as Norm. std, Deviation-Sum and PredClassMargin.

\subsection{Single vs. 2-step verbalization}
It has been reported by \citet{tian2023just} that the 2-step vs. single step verbalized numerical confidence prompts are subject to distributional shifts in their calibration of their uncertainty. To investigate this potential effect in our context, we compare the binary accuracy of the truthfulness scores, the calibration performances of the uncertainty scores and the distributions of uncertainty scores for a single vs. 2-step verbalized confidence prompt. 

\begin{table}[!ht]
    \caption{\textbf{Single vs. 2-step verbalization}}
    \centering
    \begin{tabular}{lcc}
         \textbf{Prompt}&  \textbf{single-step}& \textbf{2step}\\
         \midrule
         Binary Accuracy&  63.94\%& \textbf{65.96}\%\\
         ECE&  0.313& \textbf{0.260}\\
         Brier Score&  0.355& \textbf{0.319}\\
 K-S Test& \multicolumn{2}{c}{$\approx$ 0}\\
 \bottomrule
    \end{tabular}
    \label{tab4}
\end{table}
While the Kolmogorov-Smirnov (K-S) test reveals the 2-Step-Uncertainty prompt's uncertainty scores distribution is shifted, Table \ref{tab4} also shows that its binary classification performance on the statement truthfulness is not only sustained, but the decreased ECE score suggests better calibration. Furthermore, we report a high prevalence of 70-90\% uncertainty scores for both prompts, which is an expected result in verbalized numerical confidence prompts among various tasks \citep{huang2023look, xiong2023can, chen2023quantifying, tian2023just} (see Appendix \ref{sec: appendix_distributional_shift} for more details about the verbalized uncertainty score distributions). A deeper error analysis suggests this 2-step verbalized uncertainty procedure attains some level of calibration: Truthfulness predictions with uncertainty scores above 50 achieve a 68.6\% binary accuracy, whereas predictions with uncertainty scores below 50 are barely above chance level (52.1\%).

\subsection{BSDetector Framework} 
\begin{figure*}[!ht]
    \centering
    \caption{BSDetector Framework}
    \includegraphics[width=1.0\linewidth]{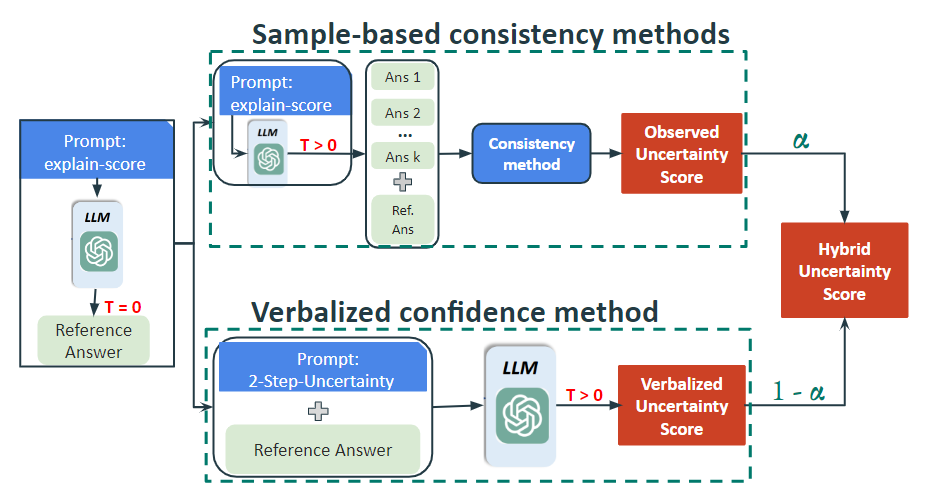}
    \label{fig3}
\end{figure*}

We have now described all sub-components that allows us to implement \citet{chen2023quantifying}'s BSDetector framework in the context of misinformation detection, as illustrated in Figure \ref{fig3}. This framework's goal is to derive a hybrid uncertainty quantification score from extrinsic (Sample-based Consistency) and intrinsic (Verbalized Confidence) uncertainty estimation methods. Specifically, we first produce a non-stochastic truthfulness score from the Explain-Score prompt; this will be our reference answer. We then produce k-stochastic sample answers from the Explain-Score prompt, which are used to derive an Observed uncertainty score \(U_{obs}\) from one of the proposed sample-based consistency methods. Furthermore, we explicitly ask the model to reflect upon its uncertainty of the reference answer and explanation via the 2-Step-Uncertainty prompt. This procedure generates a Verbalized uncertainty score \(U_{verb}\). Finally, we attain a hybrid uncertainty score by combining both scores as follows: 
\[U_{\text{hybrid}} = \alpha U_{\text{obs}} + (1 - \alpha)U_{\text{verb}}\]

where $\alpha$ is a trade-off parameter, for which we used 4-fold cross validation to hyperparamater search the optimal $\alpha$ value for each proposed method. 

\begin{figure} [!ht]
    \centering
    \caption{\textbf{Calibration curve for BSDetectork on SampleAvgDev}}
    \includegraphics[width=1\linewidth]{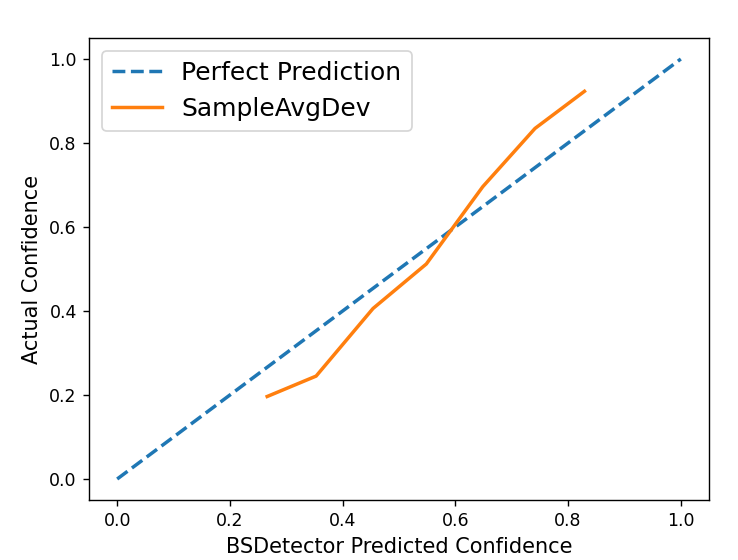}
    \label{fig4}
\end{figure}

Aligned with previous findings \cite{chen2023quantifying, xiong2023can}, the results illustrated in Table \ref{tab5} support the claim that hybrid methods largely outperform sample-based and verbalized methods. For the ECE score, we find that every proposed sample-based consistency method is improved significantly. 
\begin{table}[!h]
    \caption{\textbf{BSDetector}}
    \centering
    \begin{tabular}{lccc} 
         \textbf{Method} &  \textbf{$\alpha$} & \textbf{ECE} & \textbf{Brier Score} \\
         \midrule     
         self-consistency&  0.4& 0.119& 0.324\\ 
         selfcheckGPT&   0.7& 0.119& 0.330\\ 
         PredClassMargin&  0.4& 0.131& \textbf{0.316}\\ 
         \textbf{SampleAvgDev} & 0.9 & \textbf{0.076}& 0.334\\
         Norm. std&  0.8& 0.112& 0.322\\ 
         Deviation-Sum&   0.6& 0.133& 0.321\\ 
         \bottomrule
\label{tab5}
    \end{tabular}
\end{table}
\\In fact, when implemented in the BSDetector framework, the proposed sample-based consistency methods have close calibration performances. 
Nevertheless, we propose SampleAvgDev as the best sample-based consistency method for several reasons. First, it has the lowest ECE score with or without the BSDetector framework. Indeed, the contribution of the two-step verbalized confidence procedure is minimal, as conveyed by its high $\alpha$ value. In addition, it is robust to temperature ablation (Table~\ref{tab3}), and in cases of limited computational resources, it is still able to maintain competitive results with a small sample size (Table~\ref{tab2}). Lastly, when implemented in the BSDetector, it generates very strong uncertainty quantification, as illustrated by this method's similarity with the perfect calibration line in Figure \ref{fig4}. Consequently, we propose this method as prime candidate for GPT-4's uncertainty quantification in the context of misinformation mitigation tasks.

\subsection{Truth Scales}
\begin{table*}[!ht]
 \caption{\textbf{Truthfulness scales}}
  \centering
  \begin{tabular}{lcccccc}
    \textbf{Scale} & \textbf{Binary Accuracy} & \textbf{ROC} & \textbf{N.Ns} & \textbf{6-point Accuracy} \\
    \midrule
    \textbf{Explain-Score} & \textbf{66.98}\% & \textbf{0.6666} & 2.96\% & \textbf{28.58\%} \\
    3way-Categorical & 40\% & 0.3753 & \textbf{2.03\%} & - \\
    Politifact & 65.50\% & 0.4979 & 5.76\% & 26.64\% \\
    \bottomrule
    \label{tab6}
  \end{tabular}
\end{table*}

\begin{table*}[!ht]
    \caption{\textbf{GPT-versions}}
    \centering
    \begin{tabular}{lcccccc}
         \textbf{Prompt}&  \multicolumn{3}{c}{\textbf{Explain-score}}&  \multicolumn{3}{c}{\textbf{CoT-Explain-Score}}\\
         \midrule
         \textbf{GPT-version}& 3.5-turbo-0613&  4-0613&  \textbf{4-0314}&  3.5-turbo-0613&  4-0613& 4-0314\\
         \midrule
         \textbf{Binary Accuracy}&  63.16\% &  57.25\% &  \textbf{66.98\%} &  53.97\% &  58.41\% & 62.53\% \\
         \textbf{ROC}& 0.6269 & 0.5841 & \textbf{0.6666} & 0.5385 &  0.5887 & 0.6230 \\
         \textbf{6-point Accuracy}& 23.83\% & 23.05\% & \textbf{28.58\%} & 21.50\% & 22.59\% & 24.14\% \\
         \textbf{N.Ns}& 1.56\% & 21.57\% & 2.96\% & 4.83\% & 20.25\% & \textbf{0.17\%} \\
        \bottomrule
    \end{tabular}
    \label{tab7}
\end{table*}

We also tested with the non-dichotomized 6-way LIAR labels. A challenge here is while we mapped the 0-100 truthfulness scores to the 6-point scale uniformly, Politifact's Truth-O-meter scale description implies a requirement for a non-uniform mapping. To account for this, we explored different truthfulness scales, and evaluated which scale should be used in the BSDetector Framework. We thus compared the performances of the Explain-Score, Politifact and 3way-Categorical prompts in Table \ref{tab6} (see Appendix \ref{sec:appendix_prompt} for a detailed description of each prompt). We see, however, that 6-way performance is quite poor with all approaches, which matches the literature \cite{pelrine2023towards}---the 6-way labels may be too subjective, thus, in all the other experiments we focused on the binary ones. In addition, we note that the 3way-Categorical prompt shows poor results.

\subsection{GPT versions}
We revisited the performance of different GPT versions on binary classification. It was previously hypothesized GPT-3.5-turbo-0613 and GPT4-0613's drop in performance were due to the Explain-Score prompt's brittleness \citep{pelrine2023towards}. However, similar drops in performance are depicted in Table \ref{tab7}, regardless of the prompt. Notably, the robustness of GPT4-0613's answers drops significantly, in parallel with an increase in the Not-Numbers percentage. Given GPT-4-0314 has the best performance, we used that version in our other experiments.

\section{Conclusion}
\label{sec:conclusion}

This study investigated various uncertainty quantification methods to enhance GPT's ability to provide reliable misinformation mitigation predictions. First, we evaluated different known sample-based consistency methods that capitalized on distinct features of stochastic samples in the context of misinformation mitigation. We demonstrated how each method benefited from high levels of randomness (temperature) and how sample size affected their performance. Then, we confirmed that our two-step confidence elicitation procedure was robust in its calibration and predictive performance. In addition, we explored the robustness of the Explain-Score prompt across different versions of GPT and truthfulness numerical scales. Most importantly, we showed how to successfully implement the BSDetector framework in this domain, which combines sample-based consistency and verbalized methods. In fact, every sample-based consistency method benefited from this hybrid approach, as all proposed methods had better model calibration than their individual counterparts. Overall, our results highlight the superiority of hybrid approaches to uncertainty quantification. Due to the generality of our results, we expect this framework to provide better LLMs model calibration to other misinformation mitigation tasks and help build safer applications of NLP solutions to combat misinformation.

\section*{Acknowledgements}

This work was partially funded by the CIFAR AI Chairs Program and by the Centre for the Study of Democratic Citizenship (CSDC). 

\section*{Author Contributions}

Mauricio Rivera led the research, experiments, and writing for this project. Jean-Fran\c{c}ois Godbout and Reihaneh Rabbany advised the project, contributing ideas and feedback. Kellin Pelrine supervised the project, providing guidance and feedback at all stages.







\bibliography{ref}
\bibliographystyle{acl_natbib}

\appendix

\section{Prompts}
\label{sec:appendix_prompt}
In this section, we will describe all developed prompts used in our experiments.
\newline \newline

For the Truth Scales experiments, we considered two categorical classification prompts. 
The first `Politifact` prompt uses the format of `Explain-Score` combined with a specification of Politifact's Truth-O-Meter, which is the scale used in the LIAR dataset, to generate a 6-point truthfulness score of a statement.

\begin{displayquote}

\textbf{Politifact-scale prompt}: 'Rate the truthfulness of the following statement: "STATEMENT". Your task is to provide a score from 0 to 5, where 0: TRUE - The statement is accurate and there’s nothing significant missing. 1: MOSTLY TRUE -The statement is accurate but needs clarification or additional information. 2: HALF TRUE – The statement is partially accurate but leaves out important details or takes things out of context. 3: MOSTLY FALSE – The statement contains an element of truth but ignores critical facts that would give a different impression. 4: FALSE – The statement is not accurate. 5: PANTS ON FIRE – The statement is not accurate and makes a ridiculous claim. Do not begin your response with a number. First, write a thorough analysis, then write a vertical bar "|", then finally state your score.'
\end{displayquote}

The second prompt denoted `3way-Categorical`, uses the format of `Explain-Score` combined with \citep{chen2023quantifying}'s Self-reflection Certainty prompt. In essence, the model is asked to label the truthfulness of a LIAR dataset statement from the options: A) Correct, B) Incorrect, C) Somewhat true/Somewhat false. 

\begin{displayquote}

\textbf{3way-Categorical prompt}: 'Rate the truthfulness of the following statement: "STATEMENT" using the following options: A) Correct B) Incorrect, C) Somewhat true/Somewhat false. Do not begin your response with a number. First, write a thorough analysis, then write a vertical bar "|", then finally state your score.'    
\end{displayquote}

To analyze the robustness of our prompt format in a single vs. 2-step confidence elicitation procedure, we devised two prompts tasking GPT to give truthfulness and uncertainty scores of a given LIAR dataset statement. 
The first prompt, denoted `Single-Step-Uncertainty`, requested GPT to give 0-100 scores for both the truthfulness and uncertainty in a single step.

\begin{displayquote}

\textbf{Single-Step-Uncertainty prompt}: 'Rate the truthfulness of the following statement: "STATEMENT”. 
   Your task is to provide a truthfulness score from 0 to 100, where 0 represents definitively false and 100 represents definitively true, but you must not state your score until you have presented a thorough analysis. First, write your analysis. Then, you are required to state your truthfulness score in squared brackets “[]”. Finally, provide a different uncertainty score, where 0 represents definitely uncertain and 100 represents definitely certain of your answer, enclosed by vertical lines “|”.'
    
\end{displayquote}

The second prompt, denoted `2-Step-Uncertainty`, prompted GPT to rate its uncertainty on a 0-100 score by analyzing its own reasoning and truthfulness score generated from the Explain-Score prompt for a given LIAR dataset statement. This 2-step confidence elicitation procedure reflects the procedure used by \citet{chen2023quantifying}'s Self-reflection Certainty prompt. 

\begin{displayquote}

\textbf{2-Step-Uncertainty prompt}: \\'Statement: "STATEMENT"\\
                            Proposed answer: "GPT4 EXPLAIN-SCORE ANSWER"\\
     Your task is to rate the uncertainty of the proposed answer on a score from 0 to 100, where 0 represents definitely uncertain and 100 represents definitely certain. Please, only answer with your score.'
    
\end{displayquote}

Finally, the prompt denoted `CoT-Explain-Score` uses the Explain-Score prompt format, but specifying that the analysis must be in CoT format to produce a 0-100 truthfulness score and an analysis formated in 'thoughts', which are useful to analyze GPT's reasoning trace.

\begin{displayquote}

\textbf{CoT-Explain-Score prompt}: 'Rate the truthfulness of the following statement: "STATEMENT".
Your task is to provide a truthfulness score from 0 to 100, where 0 represents definitively false and 100 represents definitively true. First, provide a Chain of Thoughts (CoT) analysis. Then, state your truthfulness score in squared brackets “[]”. '
    
\end{displayquote}

\section{Evaluation Metrics}
\label{sec:appendix_metrics}
Here, we provide details of evaluation metrics that were not specified in the Methodology.
\newline

\textbf{6-point Accuracy}: Since the LIAR dataset uses a 6-way truthfulness classification scale, we mapped all 0-100 truthfulness scores uniformly onto a 6-point scale. Then, we denote the `6-point Accuracy` as the proportion of correctly classified truthfulness scores on this 6-point scale.

\textbf{Area under the ROC Curve (AUC) }: This 0 to 1 score provides a measure of the model's ability to distinguish classes. For instance, In our context, the higher the AUC, the better the model is at distinguishing between true and false truthfulness labels.

\section{Single vs. 2-step confidence elicitation Distributional Shift}
\label{sec: appendix_distributional_shift}
Here, we illustrate the distributional shift of our single vs. 2-step confidence elicitation procedure. Precisely, we compare the distributions of the 0-100 uncertainty scores, (scaled to 0-1 range) generated from the `Single-Step-Uncertainty` and `2-Step-Uncertainty` prompts

\begin{figure} [!ht]
    \centering
    \caption{\textbf{Distributional Shift}}
    \includegraphics[width=1\linewidth]{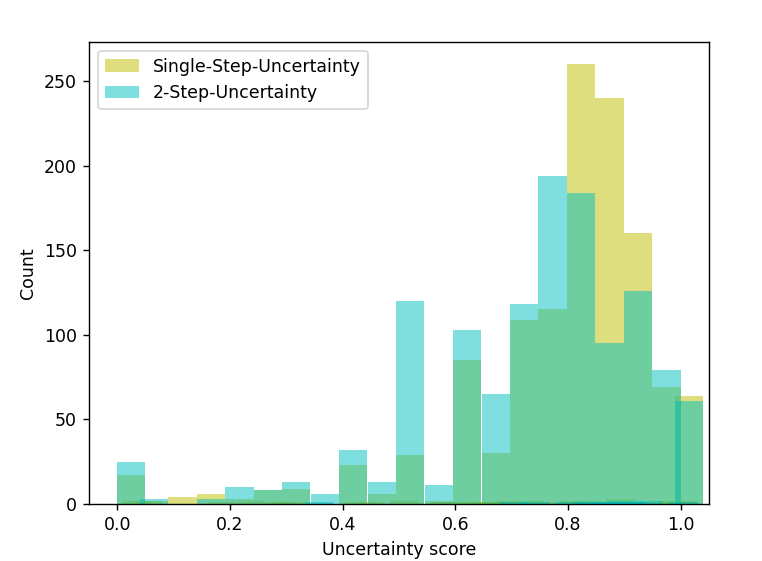}
    \label{fig5}
\end{figure}

\section{Uncertainty scores distributions of Sample-based consistency methods}
\label{sec: appendix_uncertainty_distr}
In this section, we illustrate the distribution of the uncertainty scores (scaled by 100 to produce 0-1 scores) produced by each proposed sample-based consistency method.

\begin{figure} [!ht]
    \centering
    \caption{\textbf{Self-consistency Uncertainty Scores Distribution}}
    \includegraphics[width=1\linewidth]{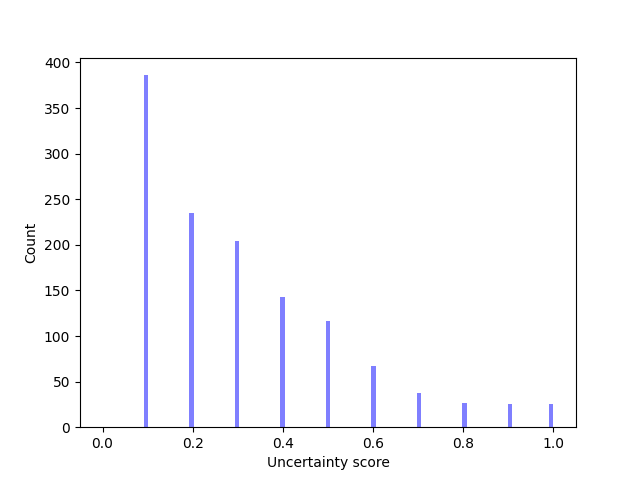}
    \label{fig6}
\end{figure}

\begin{figure} [!ht]
    \centering
    \caption{\textbf{SelfcheckGPT Uncertainty Scores Distribution}}
    \includegraphics[width=1\linewidth]{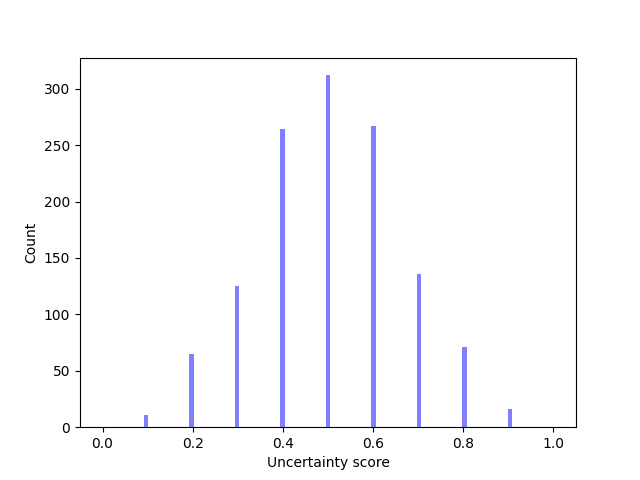}
    \label{fig7}
\end{figure}

\begin{figure} [!ht]
    \centering
    \caption{\textbf{PredClassMargin Uncertainty Scores Distribution}}
    \includegraphics[width=1\linewidth]{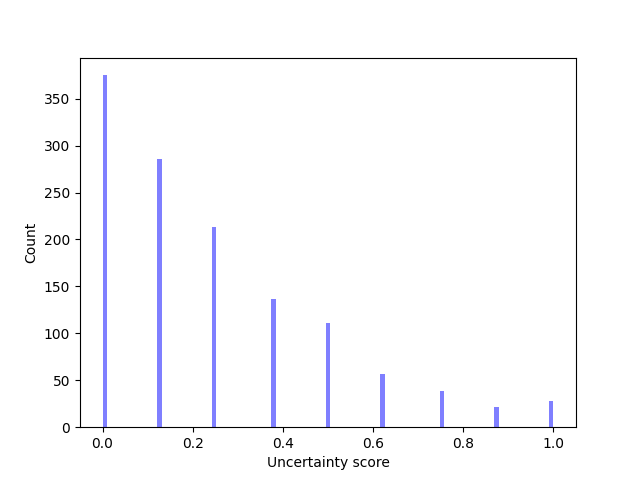}
    \label{fig8}
\end{figure}

\begin{figure} [!ht]
    \centering
    \caption{\textbf{SampleAvgDev Uncertainty Scores Distribution}}
    \includegraphics[width=1\linewidth]{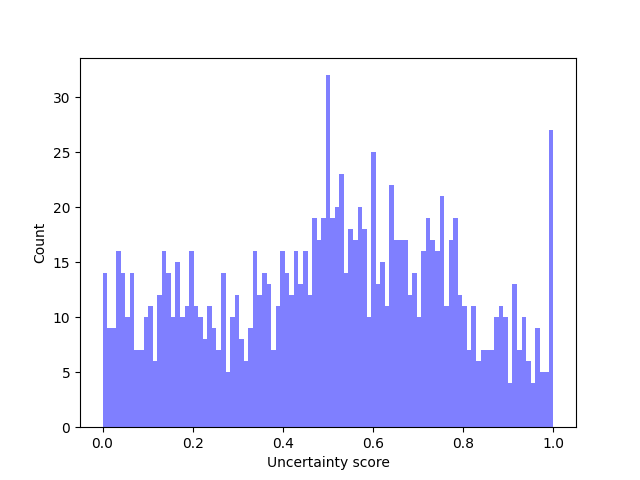}
    \label{fig9}
\end{figure}

\begin{figure} [!ht]
    \centering
    \caption{\textbf{Norm. std Uncertainty Scores Distribution}}
    \includegraphics[width=1\linewidth]{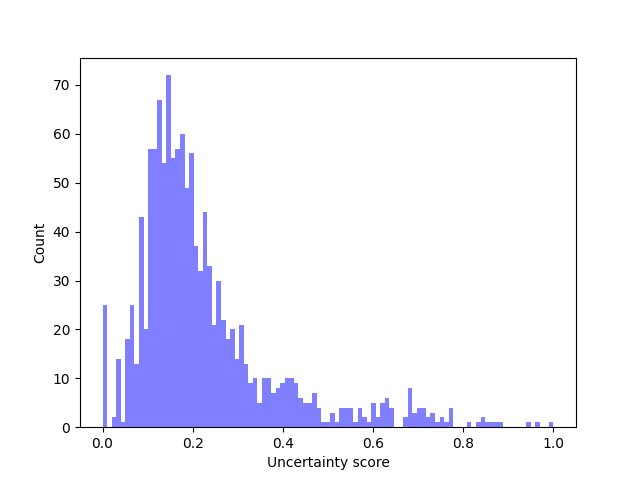}
    \label{fig10}
\end{figure}

\begin{figure} [!ht]
    \centering
    \caption{\textbf{Deviation-Sum Uncertainty Scores Distribution}}
    \includegraphics[width=1\linewidth]{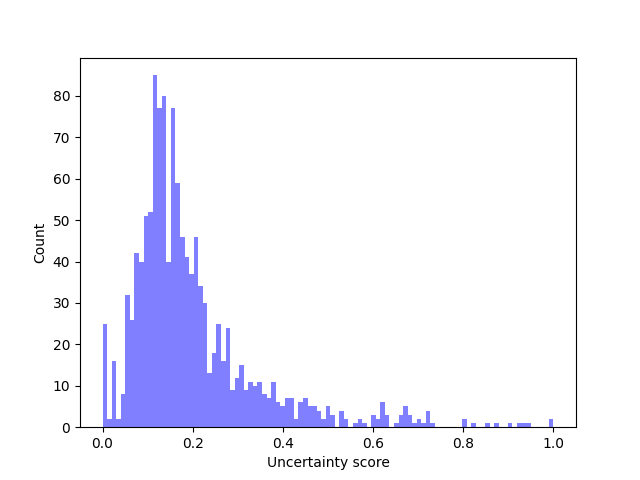}
    \label{fig11}
\end{figure}

\end{document}